\documentclass[runningheads]{llncs}
\usepackage{graphicx}
\usepackage{times}
\usepackage{soul}
\usepackage{url}
\usepackage[hidelinks]{hyperref}
\usepackage[utf8]{inputenc}
\usepackage[small]{caption}
\usepackage{amsmath}
\usepackage{amsthm}
\usepackage{booktabs}
\usepackage{algorithm}
\usepackage{algorithmic}
\usepackage{comment}
\usepackage[misc]{ifsym}
\usepackage{natbib}
\setcitestyle{round,aysep={},yysep={;}}
\usepackage{mathptmx}      % use Times fonts if available on your TeX system

\newcommand{\litnot}{\neg}
\newcommand{\sym}{\operatorname{sym}}
\newcommand{\head}{\operatorname{head}}
\newcommand{\body}{\operatorname{body}}

\newcommand{\ps}[1]{\texttt{#1}}

\newcommand{\nme}{\mathcal{E}} % Set of Near Miss Explanations

\newcommand{\vmapping}[2]{\ensuremath{\operatorname{\mathcal{V}_{#1 \mapsto #2}}}}
\newcommand{\bimapping}[2]{\ensuremath{\operatorname{{#1 \leftrightarrow #2}}}}

\newcommand{\nmx}[1]{\ensuremath{\Delta #1}}

\begin{document}
\title{Generating Contrastive Explanations for Inductive Logic Programming Based on a Near Miss Approach\thanks{Part of the work reported in this paper is funded by the Deutsche Forschungsgemeinschaft (DFG, German Research Foundation), project  427404493 (Dare2Del).
The authors declare that there is no conflict of interest. Data, material, and code used for this work can be obtained by writing the authors an e-mail. All authors contributed equally to this work.}
}
\titlerunning{Contrastive Explanations for Inductive Logic Programming}
\author{Johannes Rabold\inst{1}\orcidID{0000-0003-0656-5881} \and
Michael Siebers\inst{1}\orcidID{0000-0003-1891-3572} \and
Ute Schmid\inst{1}\orcidID{0000-0002-1301-0326}}
\authorrunning{J. Rabold et al.}
\institute{$^1$Cognitive Systems, University of Bamberg\\
\email{\{johannes.rabold,michael.siebers,ute.schmid\}@uni-bamberg.de}}
\maketitle              % typeset the header of the contribution
\begin{abstract}
In recent research, human-understandable explanations of machine learning models have received a lot of attention. Often explanations are given in form of model simplifications or visualizations. However, as shown in cognitive science as well as in early AI research, concept understanding can also be improved by the alignment of a given instance for a concept with a similar counterexample. Contrasting a given instance with a structurally similar example which does not belong to the concept highlights what characteristics are necessary for concept membership. Such near misses have been proposed by Winston (1970) as efficient guidance for learning in relational domains. We introduce an explanation generation algorithm for relational concepts learned with Inductive Logic Programming (\textsc{GeNME}). The algorithm identifies near miss examples from a given set of instances and ranks these examples by their degree of closeness to a specific positive instance. A modified rule which covers the near miss but not the original instance is given as an explanation. We illustrate \textsc{GeNME} with the well known family domain consisting of kinship relations, the visual relational Winston arches domain and a real-world domain dealing with file management. We also present a psychological experiment comparing human preferences of rule-based, example-based, and near miss explanations in the family and the arches domains.

\end{abstract}

\section{Introduction} 

Explaining classifier decisions has gained much attention in current research. If explanations are intended for the end-user, their main function is to make the human comprehend how the system reached a decision \citep{Miller19}. In the last years a variety of approaches to explainability has been proposed \citep{adadi2018peeking,molnar2019interpretable}: Explanations can be local -- focusing on the current class decision -- or global -- covering the learned model \citep{ribeiro2016should,adadi2018peeking}. A major branch of research addresses explanations by visualizations for end-to-end image classification \citep{samek2017explainable,ribeiro2016should}. Alternatively, explanations can be in form of symbolic rules \citep{lakkaraju2016interpretable,muggleton2018ultra} or in natural language \citep{stickel1991prolog,ehsan2018rationalization,SiebersS19}. A third approach to explanations is to offer prototypical examples to illustrate a model \citep{bien2011prototype,gurumoorthy2019efficient}. Finally, counterexamples can be used as counterfactuals or contrastive explanations. Counterfactuals typically are minimal changes in feature values which would have resulted in a different decision, such as: \emph{You were denied a loan because your annual income was £30,000. If your income had been £45,000, you would have been offered a loan.} \citep{wachter2017counterfactual}. In philosophy, counterfactuals have been characterized by the concept of a `closest possible world', that is, the smallest change required to obtain a different (and more desirable) outcome \citep{pollock1976possible}. Contrastive explanations have been proposed mainly for image classification. For instance, the contrastive explanation method CEM \citep{dhurandhar2018explanations} highlights what is minimally but critically absent in an image in order to belong to a given class. The MMD-critic \citep{kim2016examples} can identify nearest prototypes and nearest miss instances in image data such as handwritten digits and in Imagenet datasets. Furthermore, an algorithm ProtoDash has been proposed to identify prototypes and criticisms for arbitrary symmetric positive definite kernels which has been applied to both tabular as well as image data. 

An approach related to counterexamples has been proposed in early AI research by Winston in the context of learning relational concepts such as arch \citep{winston1970learning}. He demonstrated that presenting near miss examples where only a small number of relational aspects is missing to make an object a member of a class result in a speed up for learning. Similarly, in cognitive science research, it has been shown that alignment of structured representations helps humans to understand and explain concepts~\citep{gentner1994structural}. Gentner and Markman found that it is easier for humans to find the differences between pairs of similar items than to find the differences between pairs of dissimilar items. For example, it is easier to explain the concept of a light bulb by contrasting it with a candle than with a cat \citep{gentner1994structural}.

Induction of relational concepts has been investigated in Inductive Logic Programming (ILP) \citep{Muggleton94}, statistical relational learning \citep{koller2007introduction}, and recently also in the context of deep learning with approaches such as RelNN \citep{kazemi2018relnn} and Differentiable Neural Computers (DNCs) \citep{graves2016hybrid}. DNCs have been demonstrated to be able to learn symbolic relational concepts such as family relations or travel routes in the London underground system. These domains are typical examples for domains where ILP approaches have been demonstrated to be highly successful \citep{muggleton2018ultra}. For DNCs, questions and answers are represented as Prolog clauses. However, in contrast to ILP, the learned models are black-box. For the family domain as well as for an isomorphic fictitious chemistry domain it has been shown, that rules learned with ILP fulfill Donald Michie's criterion of ultra-strong  machine learning \citep{muggleton2018ultra}. Ultra-strong machine learning according to Michie requires a machine learning approach to teach the learned model to a human, whose performance is consequently increased to a level beyond that of the human studying the training data alone. In \citet{muggleton2018ultra} this characteristics has been related to the comprehensibility of learned rules or explanations: Comprehensibility has been defined such that a human who is presented with this information is able to classify new instances of the given domain correctly. 

\begin{figure}[t]
	% graphic with grandfather(ian, kate), grandmother(jodie, kate), grandson(ian,mat)
	\centering
	\includegraphics[width=\textwidth]{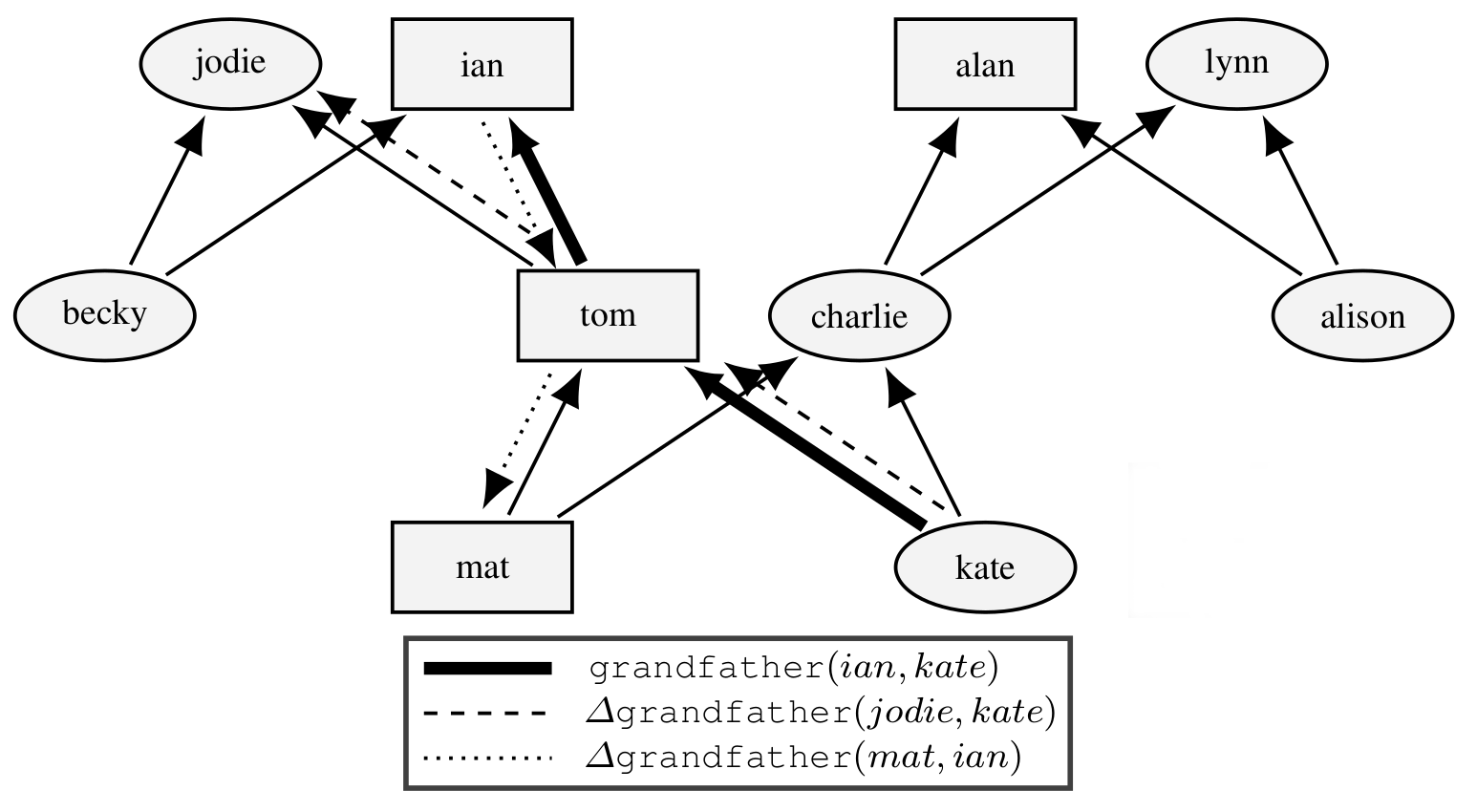}
	\caption{An example family tree. Rectangles denote male persons, ellipses denote female persons, and solid arrows denote the parent relation. The bold solid arrows indicate a trace for a positive example. Non-solid arrows indicate near miss explanations.}
	\label{fig:familytree}
\end{figure}

For ILP as well as for other relational learners such as DNCs, verbal explanations can be helpful to make a system decision transparent and comprehensible. For example, it can be explained why $\ps{grandfather}(ian, kate)$ holds by presenting the relations on the path from $ian$ to $kate$ in the family tree given in Figure~\ref{fig:familytree}: \emph{Ian is a grandfather of Kate because Ian is male and Ian is the father of Tom and Tom is the father of Kate.} Alternatively, it might be helpful for understanding the concept $\ps{grandfather}$ to present a contrastive example in form of a near miss explanation. For instance, \emph{Jodie is NOT the grandfather of Kate because she is NOT male} or \emph{Mat is NOT the grandfather of Ian because he is in a child-of-child relation to Ian and NOT in a parent-of-parent relation.} The first near miss corresponds to the concept of a \emph{grandmother}, emphasizing the importance of the attribute \emph{male} for $\ps{grandfather}$. The second near miss corresponds to the concept of a \emph{grandson}, emphasizing the importance of the relation \emph{parent}.
To our knowledge, generating such near miss examples to explain learned \emph{relational} concepts has not been investigated yet -- neither in the context of ILP nor for other machine learning approaches. 

%In contrast to counterfactuals and contrastive visual explanations as generated by CEM, near miss examples are no made-up constructions but are founded in a given domain as illustrated for \emph{grandmother} and \emph{grandson} in Figure~\ref{fig:familytree}.

In the following, we discuss the function of near miss examples. Afterwards, we present an algorithmic approach to generate near miss examples in the context of ILP and demonstrate the approach for a generic family domain, a visual domain and a real world domain dealing with file management~\citep{SiebersS19}. Finally, we present an empirical evaluation with human participants where we compare human preferences of different types of explanations for the family and the arch domain, namely rule-based global explanations, example-based explanations, as well as near miss and far miss contrastive explanations.
 
%afterwards show an application to a real world example in the context of learning irrelevancy of digital objects \citep{SiebersS19}. %TODO add again after blind review.

\section{The Function of Near Miss Examples}\label{sec:manifoldness}

\begin{figure}[t]
	% graphic with learn/explain
	\centering
        \includegraphics[width=.8\textwidth]{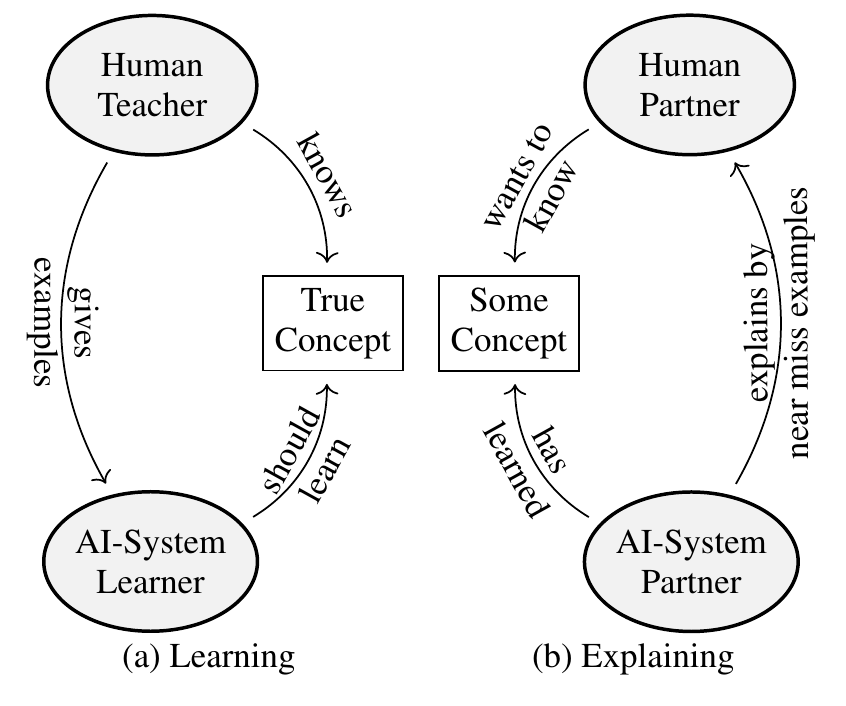}
	\caption{Duality of learning and explaining}
	\label{fig:learnexplain}
\end{figure}

Near miss examples have been introduced by Winston as a human-like strategy to machine learning \citep{winston1970learning}: A near miss example for a concept is an example which does not belong to the concept but has a strong overlap to positive examples. Such near miss examples are helpful to guide the model construction of a machine learning algorithm \citep{telle2019teaching}. Winston illustrated learning with near misses in the context of relational visual domains. Concepts are represented as compounds of primitive blocks such as cubes. For instance, positive examples for arches must consist of at least two objects playing the role of supporters (pillars) and another object on top of them (roof). Negative examples for an arch might be a tower of several cubes -- a far miss -- or two pillars with no space between them covered by a roof -- a near miss.

In the context of machine learning of relational concepts such as  Winston's arches, molecules \citep{king1996structure}, or Turing-complete languages \citep{telle2019teaching}, carefully constructed near misses given as negative examples can speed up learning considerably (see Figure~\ref{fig:learnexplain}.a). In this case, the machine learning expert plays a role similar to that of a school teacher who identifies helpful examples \citep{schmid2003closer}. We propose that what is effective for learning is also effective for explaining a learned model (see Figure~\ref{fig:learnexplain}.b): for some concept that an AI system has learned, it can explain its model by constructing a near miss example. While machine learning typically is unidirectional -- the human provides the training examples and the system generalizes a model -- explanations can support interactive machine learning scenarios based on a bidirectional partnership between human and AI system \citep{nguyen2018believe}.

While there are some considerations about what constitutes helpful examples in educational psychology \citep{gentner2003learning} and the insights given in Winston's seminar work \citep{winston1970learning}, there exist no general principles to construct helpful near miss explanations. We base our algorithm presented in the next section on some general observations which we will illustrate with the family domain example of Figure~\ref{fig:familytree}.

\section{Near Miss Explanations}

In the following, we will introduce the \textsc{GeNME} algorithm for generating near miss explanations. Our approach extends the comprehensible machine learning approach ILP~\citep{muggleton2018ultra} with a contrastive explanation component. First, we will introduce the basic notation and basic concepts of ILP. The concept of a near miss explanation is introduced formally and the generation algorithm is presented.

\subsection{Notation}\label{sec:notation}
We introduce basic notation for first-order logic theories based on the clause form underlying the logic programming language Prolog~\citep{sterling1994art}.
Following Prolog's notational conventions,
% variables, functions, predicates
variables, constants and predicate symbols are represented as strings of letters, numbers, and underscores where variables must start with an upper case letter and constants and predicate symbols with a lower case letter. The arity of a predicate symbol is the number of arguments it takes.
A predicate is called \emph{attribute} if it has arity one and \emph{relation} otherwise.

% atoms, literals
Every variable as well as every constant is a \emph{term}. We call constants \emph{ground terms}.

A predicate symbol of arity~$n$ followed by a bracketed $n$-tuple of terms is called \emph{atom}, or \emph{positive literal}. Function $\sym(A)$ returns the predicate symbol of atom $A$. The negation of an atom is called \emph{negative literal}. The negation symbol is $\litnot$. A literal is \emph{ground} if all terms in its $n$-tuple are \emph{ground}. 

% clause, theory
A \emph{clause} is an implication where the antecedent is a set of literals and the consequent is an atom. We write the implication reversed, as $H \gets \lbrace L_1, \ldots, L_m \rbrace$. 
The consequent of a clause~$C$ is called its \emph{head}, $\head(C)$. Its antecedent is called the \emph{body} of the clause, $\body(C)$.
For convenience, we may omit the braces surrounding the body.
If the body of clause~$C$ is the empty set, $H \gets \lbrace\rbrace$, we call $C$ a \emph{fact}, omit the antecedents, and simply write $H$. 
A clause is called ground when all its literals are ground. A set of clauses is called a \emph{(clausal) theory}.

% substitution
A substitution is a mapping from variables to terms. We denote a substitution $\theta$ by $\lbrace x_1 \mapsto t_1, \ldots, x_k \mapsto t_k \rbrace$ where $x_1, \ldots, x_k$ are variables and $t_1, \ldots, t_k$ are terms. 
A substitution is applied to a term by simultaneously replacing all $x_i$ in the term by the corresponding $t_i$'s. A substitution is applied to a literal by applying it to all terms in the literal. A substitution is applied to a clause by applying it to all literals in the clause. We denote the application of the substitution~$\theta$ to a term, literal, or clause~$X$ by $X\theta$.
% truth values
If a literal or a set of literals~$K$ is true given a clausal theory~$T$, we say that $T$ models $K$, or $T \models K$.
Theory~$T$ models an atom~$A$ if there exists a clause~$C \in T$ and a substitution~$\theta$ such that $A=\head(C\theta)$ and $T \models \body(C\theta)$.
Using negation by failure, a clausal theory~$T$ models a negative literal $\litnot A$ if $T$ does not model $A$, $T \not\models A$. A theory~$T$ models a set of literals~$\lbrace L_1, \ldots, L_n \rbrace$ if there is a substitution~$\theta$ such that $T$ models $L_i\theta$ for $1\leq i \leq n$.
By definition, the empty set~$\lbrace \rbrace$ is modeled by any theory.

\subsection{Basic Concepts of ILP}\label{sec:ilp}

\begin{figure}[t]
	\centering
	\textbf{Background Knowledge:}\\
	\begin{tabular}{ll}
		\texttt{female(jodie)} & \texttt{parent(jodie, becky)}\\
		\texttt{female(lynn)} & \texttt{parent(jodie, tom)}\\
		\texttt{female(becky)} & \texttt{parent(ian, becky)} \\
		\texttt{female(charlie)} & \texttt{parent(ian, tom)}\\
		\texttt{female(alison)} & \texttt{parent(alan, charlie)}\\
		\texttt{female(kate)} & \texttt{parent(alan, alison)}\\
		\texttt{male(ian)} & \texttt{parent(lynn, charlie)}\\
		\texttt{male(alan)} & \texttt{parent(lynn, alison)} \\
		\texttt{male(alan)} & \texttt{parent(tom, mat)} \\
		\texttt{male(tom)} & \texttt{parent(tom, kate)} \\
		\texttt{male(mat)} & \texttt{parent(charlie, mat)} \\
		& \texttt{parent(charlie, kate)}
	\end{tabular}
	
	\textbf{Selection of positive examples:}\\
	\texttt{grandfather(ian, kate)}\\
	\texttt{grandfather(alan, mat)}
	
	\textbf{Selection of negative examples:}\\
	\texttt{grandfather(alan, tom)}\\
	\texttt{grandfather(jodie, mat)}
	\caption{Background knowledge for the family domain together with a selection of positive and negative examples for the \ps{grandfather} concept.}
	\label{fig:family_ilp}
\end{figure}

ILP is a sub-field of symbolic machine learning which deals with learning clausal theories from examples~\citep{Muggleton94}. Such clausal theories allow to represent relational concepts where either the target defines a relation (i.e. has more than one argument) or the target is defined over relational structures such as Winston's arches. For instance, a theory for \ps{grandfather} can be learned from positive examples, such as $\ps{grandfather}(ian, kate)$, and negative examples, such as\\ $\ps{grandfather}(alan, tom)$ (see Figure~\ref{fig:family_ilp}). Positive and negative examples for the target concept are represented as ground atoms. Additionally, a background knowledge theory is provided. In the family domain, the facts $\ps{parent}(tom, kate)$ and $\ps{male}(ian)$ can be part of the background knowledge. The learned theory together with the background knowledge theory must model all positive examples and no negative examples.

Assume the learned theory for $\ps{grandfather}$ consisting of a single clause,
\begin{equation}
\begin{aligned}
\ps{grandfather}(A,B) \gets & \ps{male}(A), \\
& \ps{parent}(A,C), \\
& \ps{parent}(C,B) \text{.} \label{eq:grandfather}
\end{aligned}
\end{equation}

\noindent In general, a learned theory can include several clauses characterizing the target concept. For example, the target concept \ps{grandparent} can be described by a set of clauses taking into account the genders of the respective parents. It can also be the case that target clauses are not exclusive. That is, a positive example $P$ might follow from multiple clauses.

With the learned theory, new instances given as ground atoms can be classified. For example, $\ps{grandfather}(alan, kate)$ will be classified as positive;\\ $\ps{grandfather}(becky, tom)$ will be classified as negative.

\subsection{Near Miss Examples and Explanations}\label{sec:prob_form}
Such positive classified instances are modeled by the learned theory as introduced in Subsection~\ref{sec:notation}. As mentioned above, theory $T$ consists of predefined background clauses and clauses learned for the target concept. For example, the \ps{grandfather} relation holds for $ian$ and $kate$ given the theory in Equation~\ref{eq:grandfather} together with background knowledge about parent relations and the gender of persons in a given family domain as the one shown in Figure~\ref{fig:family_ilp}. An explanation for this fact has to make explicit how this can be derived from the theory. The reason why $\ps{grandfather}(ian, kate)$ holds is that $ian$ is male and $ian$ is a parent of $tom$ and $tom$ is a parent of $kate$. In general, we call an explanation for a positive example \emph{local explanation}:

\begin{definition}[Local Explanation]\label{def:local_explanation}
	A local explanation for a positive example $P$ is a ground clause $C\theta$ where $C \in T$ such that $P = \head(C\theta)$ and $T \models \body(C\theta)$.
\end{definition}

To emphasize which information is crucial for making someone a grandfather of someone else, a near miss explanation might be helpful. As argued in Section~\ref{sec:manifoldness}, a near miss example is not a positive instance for the target concept, but illustrates a semantically similar concept. For example given the positive example\\$\ps{grandfather}(ian, kate)$, possible near miss examples could be the female parent of a parent (that is the grandmother) of $kate$ or a male child of a child (that is a grandson) of $ian$.
Formally, we define near miss explanations and near miss examples:

\begin{definition}[Near Miss Explanation]\label{def:nearmiss}	
	Given a local explanation $C\theta$ and a minimally changed clause $C^\prime$ with substitution $\theta$, we call $C^\prime\theta^\prime$ a near miss explanation and $\nmx{\head(C^\prime\theta^\prime)}$ a near miss example if $T \models \body(C^\prime\theta^\prime)$, $T \not\models \head(C^\prime\theta^\prime)$. $\nmx$ is marking literals as near miss examples.
\end{definition}

\begin{comment}
\begin{definition}[Near Miss Explanation]\label{def:nearmiss}
Given a miss example $\mx{\head(C^\prime\theta^\prime)}$, we call $\nmx{\head(C^\prime\theta^\prime)}$ a near miss example if there is no other miss example $\mx{\head(C^\prime\theta^{\prime\prime})}$ such that $\theta^{\prime\prime} <_\theta \theta^\prime$.
Accordingly, we call the corresponding miss explanation $E_P = \body(C^\prime\theta^\prime)$ a near miss explanation. 
\end{definition}
\end{comment}

What constitutes a minimal changed clause is domain dependent. In general, we understand changing a clause as replacing literals in its body by different literals. The most basic change is replacing a single literal by its negation. For example, the attribute \ps{male} could be changed to $\litnot \ps{male}$; the relation \ps{parent} could be changed to $\litnot \ps{parent}$. In a geometric domain, an attribute \ps{large} could be changed to $\litnot \ps{large}$; a relation \ps{above} could be changed to $\litnot \ps{above}$. However, such negations are too unspecific for many domains. In a more fine-grained geometry domain, $\litnot \ps{large}$ could mean \ps{small} or \ps{medium}. Therefore, we propose that it is helpful to define pairs of \emph{semantically opposing} predicate symbols, for example inverses, in $T$ when modeling a particular domain. In natural language semantics, such relational opposites are one of the basic relations between lexical units~\citep{palmerrobert}.

In the family domain, the pairs \ps{male}/\ps{female} and \ps{parent}/\ps{child} are semantic opponents. To explain $\ps{grandfather}(ian, kate)$, the near miss example \\$\nmx{\ps{grandfather}(jodie, kate)}$ (which is actually the grandmother) can be derived by replacing $\ps{male}(A)$ with $\ps{female}(A)$ in Equation~\ref{eq:grandfather}.
An alternative near miss can be derived by inverting the \ps{parent} relation to \ps{child}. Because \ps{grandfather} relies on the transitive sequence of two \ps{parent} relations, both occurrences should be replaced, resulting in $\nmx{\ps{grandfather}(mat, ian)}$ (which should actually read $\ps{grandson}(mat, ian)$).

Depending on the domain, a minimal change of a clause $C$ might therefore consist of either replacing a single literal or multiple literals. Which literals may be replaced may also depend on the semantic opponents involved. Thus, we introduce domain dependent rewriting filters \vmapping{p}{q} to formalize this connection. $\vmapping{p}{q}(B)$ extracts all valid literal sets from clause body $B$, such that the predicate symbol $p$ may be replaced by predicate symbol $q$ in those literal sets. For example, \vmapping{\ps{male}}{\ps{female}} applied to the body of Equation~\ref{eq:grandfather} yields $\big\lbrace \lbrace \ps{male}(A) \rbrace \big\rbrace$ and \vmapping{\ps{parent}}{\ps{child}} applied to the body of Equation~\ref{eq:grandfather} yields \\$\big\lbrace \lbrace \ps{parent}(A, C), \ps{parent}(C,B) \rbrace \big\rbrace$. That is, either \ps{male} may be replaced (by \ps{female}) or both \ps{parent}'s (by \ps{child}). These rewriting filters can be selected by the respective user of the algorithm.

As the head of a clause is never changed, $\head(C)=\head(C^\prime)$, $\theta^\prime$ must be different from $\theta$ for each near miss explanation $C^\prime\theta^\prime$ w.r.t.\@ local explanation $C\theta$. Otherwise, the near miss example would be identical to the positive example, $\head(C^\prime\theta^\prime)=\head(C\theta)=P$, which is by definition not a near miss example ($T \models P$).
Based on this change we define the degree of a near miss explanation:

\begin{definition}[Degree of Near Miss Explanation]\label{def:degree}
	Given a near miss explanation $C^\prime\theta^\prime$ w.r.t.\@ local explanation $C\theta$, the degree $n$ of the near miss explanation is the number of changed replacements, $n = \lvert \theta \setminus \theta^\prime \rvert$.
\end{definition}

In our approach, we derive near miss examples rather than counterfactuals. That is, near misses must be defined over constants in the background knowledge. It is not possible to invent new constants. 
Thus, we can define possible near miss examples, that is \emph{near miss candidates}:

\begin{definition}[Near miss candidate]\label{def:candidate}
	A \emph{near miss candidate} for a positive example $P$ is a ground atom~$N$ which has the same predicate symbol as $P$, $\sym(N) = \sym(P)$, but is not modeled by the theory $T$, $T \not\models N$.
\end{definition}

\begin{algorithm}[t]
	\caption{\textsc{GeNME}: The Near Miss Explanation Generation Algorithm}
	\label{alg:near_miss_algorithm}
	\begin{algorithmic}[1]
		\REQUIRE~Positive Example $P$, Theory $T$, Set of rewriting filters $O$ 
		\STATE Initialize family of result sets $(\nme_i)$ to empty sets
		
		\STATE $\mathcal{N} \gets \lbrace N~\mid~\sym(N) = \sym(P), T \not\models N \rbrace$
		
		\FORALL{$C \in T$ and local explanations $C\theta$ for $P$}\label{genme.local_exps}

		\FORALL{$\vmapping{p}{q} \in O$}\label{genme.filters}
		\FORALL{$\mathcal{L} \in \vmapping{p}{q}(\body(C))$}\label{genme.literalsets}
		
		\STATE $C^\prime \gets \big( \head(C) \gets \left( \body(C) \setminus\mathcal{L} \cup \ps{rename}(\mathcal{L}, p, q) \right) \big)$\label{genme.candidate_nme}
		
		\FORALL{$N \in \mathcal{N}$}\label{genme.for_candidates}
		
		\STATE $d \gets 0$
		\STATE $\mathcal{E}^\prime \gets \lbrace \rbrace$
		
		\WHILE{$\mathcal{E}^\prime = \lbrace \rbrace$ and $d < \lvert \theta \rvert$}\label{genme.while}
		
		\STATE{$d \gets d+1$}
		
		\FORALL{$\theta^\prime$ such that $\lvert \theta \setminus \theta^\prime \rvert = d$}\label{genme.theta_prime}
		\STATE $\mathcal{E}^\prime \gets \mathcal{E}^\prime \cup \lbrace C^\prime\theta^\prime \mid T \models \body(C^\prime\theta^\prime)$ and $\head(C^\prime\theta^\prime) = N \rbrace$\label{genme.test}
		\ENDFOR
		\ENDWHILE
		
		\STATE $\mathcal{E}_d \gets \mathcal{E}_d \cup \mathcal{E}^\prime$\label{genme.add}
		
		\ENDFOR
		
		\ENDFOR
		\ENDFOR
		\ENDFOR
		
		\RETURN $(\nme_i)$
	\end{algorithmic}
\end{algorithm}
\begin{algorithm}[t]
	\caption{\ps{rename}: Replace predicate symbol in a set of literals by opponent}
	\label{alg:rename}
	\begin{algorithmic}[1]
		\REQUIRE~Set of literals $\mathcal{L}$, Predicate symbols $p$, $o$
		\STATE $\mathcal{L}^\prime \gets \lbrace\rbrace$
		\FORALL{$L \in \mathcal{L}$}
		\IF{$L = p(a_1, \dotsc, a_n)$}
		\STATE $\mathcal{L}^\prime \gets \mathcal{L}^\prime \cup \lbrace o(a_1, \dotsc, a_n) \rbrace$
		\ELSIF{$L = \litnot p(a_1, \dotsc, a_n)$}
		\STATE $\mathcal{L}^\prime \gets \mathcal{L}^\prime \cup \lbrace \litnot o(a_1, \dotsc, a_n)\rbrace$
		\ENDIF
		\ENDFOR
		\RETURN $\mathcal{L}^\prime$
	\end{algorithmic}
\end{algorithm}

\subsection{Algorithm}

To generate near miss explanations we introduce \textsc{GeNME} (Algorithm~\ref{alg:near_miss_algorithm}). Given a positive example $P$ represented as a ground atom, a domain theory $T$ given as a set of clauses modeling the target concept together with clauses that describe the application domain, and a set of rewriting filters $O$, \textsc{GeNME} returns a set of sets of near miss examples $(\mathcal{E}_d)$. Each element of $(\mathcal{E})$ contains near miss explanations of a given degree~$d$, inducing a partial ordering over counterexamples in relation to instance $P$.

To generate all near miss examples for the given positive example, \textsc{GeNME} first generates the set of all near miss candidates and then iterates over all local explanations. To make sure that our algorithm terminates, we only allow near miss candidates that are ground with constants already present in the domain.
For each local explanation, we iterate over all valid literal sets (line~\ref{genme.filters} and \ref{genme.literalsets}) to generate a minimally changed clause $C^\prime$ by renaming predicate symbol $p$ to $o$ (line~\ref{genme.candidate_nme} applying Algorithm~\ref{alg:rename}).

For each such minimally changed clause~$C^\prime$, \textsc{GeNME} iterates over all near miss candidates (line~\ref{genme.for_candidates}) and all possible substitutions $\theta^\prime$ in increasing distance from $\theta$ (lines~\ref{genme.while}--\ref{genme.theta_prime}). If there are substitutions $\theta^\prime$ such that $\head(C^\prime\theta^\prime)$ equals the near miss candidate and the theory models $\body(C^\prime\theta^\prime)$ for a given distance (line~\ref{genme.test})
we add all near miss explanations for this distance to $\mathcal{E}_d$ (lines~\ref{genme.test} and \ref{genme.add}). As soon as one near miss explanation $\mathcal{E}'$ for the given $d$ is found, we continue with the next near miss candidate $N \in \mathcal{N}$.

\subsection{Termination, Time Complexity and Implementation Details}

The four nested \textbf{for all} loops in \textsc{GeNME} (starting in lines~\ref{genme.local_exps}, \ref{genme.filters}, \ref{genme.literalsets} and \ref{genme.for_candidates}) all iterate over finite sets and therefore are guaranteed to terminate. The \textbf{while} loop (starting in line~\ref{genme.while}) with the included \textbf{for all} loop is eventually terminated when we reach a degree that coincides with the magnitude of the substitution $\vert \theta \vert$.

\begin{comment}
\textsc{GeNME} terminates for all admissible inputs given a finite theory $T$ and a finite set of rewriting filters $O$ since given that constraint, there can be no infinite loop when iterating over all $C \in T$ and all $\vmapping{p}{q} \in O$. Additionally, when iterating over all near miss candidates $N \in \mathcal{N}$, the loop is exited when there are no more candidates. While searching for a near miss explanation for a given degree $d$, we stop searching if we reach a degree that coincides with the magnitude of the substitution $\vert \theta \vert$. When we assume clauses with a finite number of literals, this loop will be exited eventually.
\end{comment}

\textsc{GeNME} is correct in the sense that it does not output explanations that are positive examples for the given theory. This is achieved by only considering rule heads that are not modeled by the theory. Also the algorithm is guaranteed to find the miss explanations with lowest degree $d$ for a given minimally changed clause $C^\prime$ since we iteratively increment $d$ and exit the search as soon as we find a miss explanation for $C^\prime$.

In its given explicit, non-optimized realization, Algorithm~\ref{alg:near_miss_algorithm} has a time complexity dependent on the number of constants $\mathcal{C}$ in the background theory. The number of $C \in T$ and the number of $\vmapping{p}{q} \in O$ has a linear impact on runtime. The upper bound for the number of $\mathcal{N}$ is given by the magnitude of the Cartesian product $\mathcal{C}^n$ where $n$ denotes the arity of the target literal. The upper bound is reached when no grounding for the target literal is modeled by theory $T$.

The algorithm iterates over $N \in \mathcal{N}$ and tests if a substitution $\theta^\prime$ can be found for the minimally changed clause $C^\prime$ such that $T \models body(C^\prime\theta^\prime)$ and $head(C^\prime\theta^\prime) = N$. The algorithm first tries to find a near miss explanation by testing a $\theta^\prime$ where only one substitution rule is different from the original $\theta$ ($d = 1$). If no near miss explanation is found, the degree $d$ is gradually incremented until a near miss explanation is found or $d$ reaches $\vert \theta \vert$. For a given $d$ the number of possible combinations of substitution rules to change in $\theta$ is given by $\binom{\vert \theta \vert}{d}$. For each rule the number of new terms to use as substitute is given by $\mathcal{C}-1$. The number of possible $\theta^\prime$ for a given $d$ is therefore given by $\binom{\vert \theta \vert}{d} \cdot (\vert \mathcal{C} \vert - 1)^d$. For the family domain the numbers would be $\vert \theta \vert = 3$ and $\vert \mathcal{C} \vert = 10$. The respective runs for all $d$'s would add up to $\sum_{d=1}^3{\binom{\vert \theta \vert}{d} \cdot (\vert \mathcal{C} \vert - 1)^d} = 999$ altered substitutions that would have to be checked.

The inefficiency of the algorithm is mainly due to its unsophisticated filtering strategy. For an implementation, this can be realized more efficiently by realizing the following considerations: Since we are only interested in misses, we can constrain the substitutions for variables occurring in the head of $C^\prime$ to the ones dictated by the current $N \in \mathcal{N}$. When we also check the equality of head$(C^\prime\theta^\prime)$ and $N$ before we check $T \models \textrm{body}(C^\prime\theta^\prime)$ we can safely continue with the next $\theta^\prime$. Additionally the user can be featured with the possibility to label particular constants in the local explanation as immutable. This way the labeled constants will not be part of the search for alternative substitutions $\theta^\prime$ which can improve overall efficiency.
An efficiency boost can also be reached by choosing rewriting filters that make sense in the chosen domain and that are also occurring in $T$.

\section{Application to Example Domains}

In the following subsections we show that \textsc{GeNME} generates plausible near miss explanations. We present the results of applying the algorithm to different domains: a generic family domain constituted by abstract family relations such as grandfather, a relational visual domain of blocksworld arches, and a real world domain dealing with file management.

\subsection{Family Domain}\label{sec:app_family}

%booktabs (toprule, midrule, bottomrule)
\begin{table}[t]
	\centering
	\caption{Number of found near miss explanations by degree in the family domain. $\mathcal{N}$ denotes the set of all near miss candidates, $\mathcal{E}_d$ the set of near miss explanations of degree $d$, \ps{gf} the \ps{grandfather} relation, and \ps{dt} the \ps{daughter} relation. $\bimapping{x}{y}$ denotes the use of $\mathcal{V}_{x \mapsto y}$ or $\mathcal{V}_{y \mapsto x}$, respectively.}
	\label{tab:family}
	\begin{tabular}{ccc}
		& $\ps{gf}(ian, kate)$ & $\ps{dt}(becky, jodie)$\\
		& $\lvert\mathcal{N}\rvert = 96$ & $\lvert\mathcal{N}\rvert = 92$\\
		\toprule
		$\bimapping{male}{female}$ &&\\
		$\lvert \mathcal{E}_1 \rvert$ & 1 & 1\\
		$\lvert \mathcal{E}_2 \rvert$ & 2 & 3\\
		$\lvert \mathcal{E}_3 \rvert$ & 1 & 0\\
		\midrule
		$\bimapping{parent}{child}$ &&\\
		$\lvert \mathcal{E}_1 \rvert$ & 0 & 0\\
		$\lvert \mathcal{E}_2 \rvert$ & 2 & 6\\
		$\lvert \mathcal{E}_3 \rvert$ & 2 & 0\\
		\bottomrule
	\end{tabular}
\end{table}

The family domain models parent-child relations between persons. This domain is represented by attributes \ps{male} and \ps{female}, and relations \ps{parent} and \ps{child}. In Figure~\ref{fig:family_ilp} the background clauses of the theory (excluding the \ps{child} relation, which is the inverse of \ps{parent}) are stated. Additionally, the theory contains the two clauses from Equations~\ref{eq:grandfather} and \ref{eq:daughter}.

\begin{equation}
\ps{daughter}(A,B) \gets \ps{female}(A), \ps{child}(A,B)
\label{eq:daughter}
\end{equation}

Given this domain, we use \textsc{GeNME} to explain the positive examples \\$P_1 = \ps{grandfather}(ian, kate)$ and $P_2 = \ps{daughter}(becky, jodie)$. We provide four rewriting filters: $\vmapping{\ps{male}}{\ps{female}}$, $\vmapping{\ps{female}}{\ps{male}}$, $\vmapping{\ps{parent}}{\ps{child}}$, and\\ $\vmapping{\ps{child}}{\ps{parent}}$, where the first two filters allow changing a single occurrence of the predicate and the second ones require changing all occurrences.

Out of the 96 candidates for $P_1$, \textsc{GeNME} identifies 8 as near miss examples (8.3\,\%, see Table~\ref{tab:family}). The near miss explanation with the lowest degree of 1 is \\$\ps{grandfather}(jodie, kate) \gets \ps{female}(jodie), \ps{parent}(jodie, tom),\\ \ps{parent}(tom, kate)$. Indeed, this is intuitively a very close near miss example (the grandmother of $kate$). Also for $P_2$, \textsc{GeNME} finds an intuitively very close near miss explanation ($jodie$'s son) as solitary near miss explanation with lowest degree, \\$\ps{daughter}(tom, jodie) \gets \ps{male}(tom), \ps{child}(tom, jodie)$. In total, \textsc{GeNME} identifies 10 out of 92 candidates for $P_2$ as near miss examples (10.9\,\%).

For both minimal near miss examples, \ps{male} was replaced by \ps{female} or vice versa. Also for the inverse relations of \ps{parent} or \ps{child}, plausible near misses were found: for instance, that $mat$ is the grandson of $ian$ and $jodie$ is the mother of $becky$. Both near miss examples have degree 2, illustrating that they are not as near as the near miss examples given above.

\subsection{The Winston Arches}\label{sec:app_arches}

\begin{figure}[t]
	\centering
	\textbf{Background Knowledge:}\\
	\begin{tabular}{ll}
		\texttt{contains(struct1, a1)} & \texttt{contains(struct4, a1)}\\
		\texttt{contains(struct1, b)} & \texttt{contains(struct4, b)}\\
		\texttt{contains(struct1, c)} & \texttt{contains(struct4, c)} \\
		\texttt{supports(b, a1, struct1)} & \texttt{supports(b, a1, struct4)}\\
		\texttt{supports(c, a1, struct1)} & \texttt{supports(c, a1, struct4)}\\
		\texttt{not\_meets(b, c, struct1)} & \texttt{meets(b, c, struct4)}\\
		\texttt{contains(struct2, a1)} & \texttt{contains(struct5, a2)}\\
		\texttt{contains(struct2, b)} & \texttt{contains(struct5, b)} \\
		\texttt{contains(struct2, c)} & \texttt{contains(struct5, c)} \\
		\texttt{supports(b, a1, struct2)} & \texttt{supports(a2, b, struct5)} \\
		\texttt{supports(c, a1, struct2)} & \texttt{supports(a2, c, struct5)} \\
		\texttt{not\_meets(b, c, struct2)} & \texttt{not\_meets(b, c, struct5)} \\
		\texttt{contains(struct3, a1)} & \texttt{contains(struct6, a2)}\\
		\texttt{contains(struct3, b)} & \texttt{contains(struct6, b)} \\
		\texttt{contains(struct3, c)} & \texttt{contains(struct6, c)} \\
		\texttt{supports(b, a1, struct3)} & \texttt{supports(b, a2, struct6)} \\
		\texttt{supports(c, a1, struct3)} & \texttt{supports(c, a2, struct6} \\
		\texttt{not\_meets(b, c, struct3)} & \texttt{meets(b, c, struct6)} \\
		\multicolumn{2}{c}{\texttt{supported\_by(X, Y, A) $\gets \neg$\texttt{supports(Y, X, A)}}}
	\end{tabular}
	\caption{Background knowledge for the Winston arches domain.}
	\label{fig:arches_ilp}
\end{figure}

\begin{figure}[t]
	\centering
	\includegraphics[width=.7\textwidth]{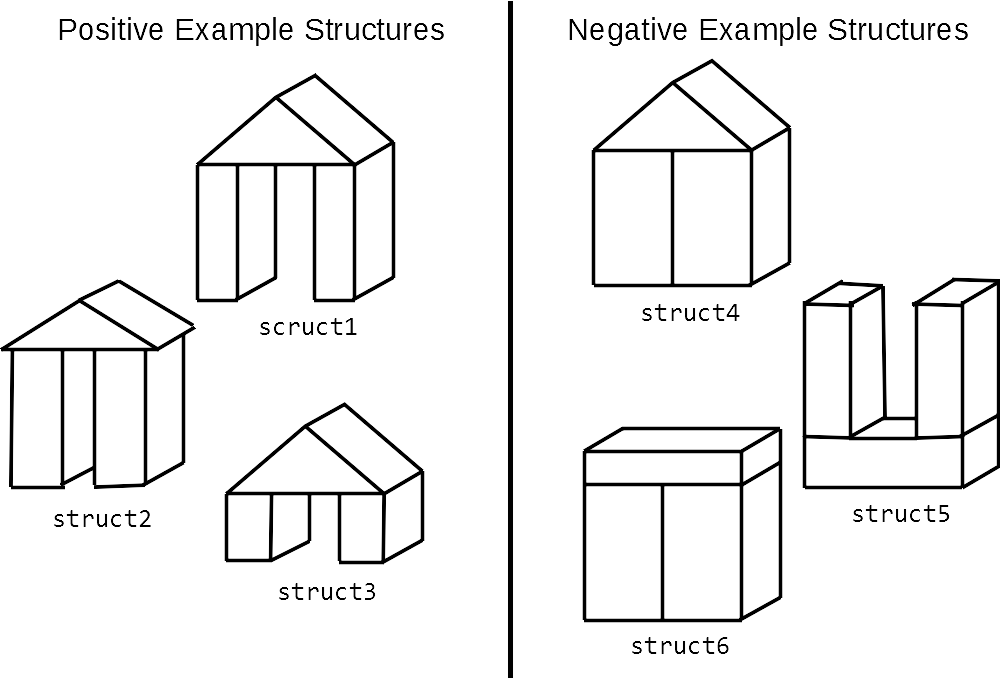}
	\caption{The positive and negative example structures for the Winston arches domain.}
	\label{fig:posneg_arches}
\end{figure}

The Winston arches visual domain models structural relations of building blocks. The domain features relations \ps{contains} (a structure contains a block), \ps{supports} (a block supports another block), \ps{(not\_)meets} (two blocks do (not) meet horizontally) and \ps{is\_a} (a block has a certain shape; the shape can either be \ps{wedge} or \ps{brick}). Figure~\ref{fig:arches_ilp} states the background clauses for the domain. Figure~\ref{fig:posneg_arches} additionally visualized the positive and negative examples for the concept \ps{arch}. We further assume the learned concept for \ps{arch} in Equation~\ref{eq:arch}.

\begin{equation}
\begin{aligned}
\ps{arch}(A) \gets & \ps{contains}(A, X), \ps{contains}(A, Y), \ps{contains}(A, Z), \\
& \ps{is\_a}(X, wedge), \ps{is\_a}(Y, brick), \ps{is\_a}(Z, brick), \\
& \ps{supports}(Y, X, A), \ps{supports}(Z, X, A), \\
& \ps{not\_meets}(Y, Z, A) \label{eq:arch}
\end{aligned}
\end{equation}

When using \textsc{GeNME} with this domain we introduced a small change for handling constants that are contained in the theory (e.g. $wedge$, $brick$ in Equation~\ref{eq:arch}). In order to find alternative substitutions $\theta^\prime$ for this constants, we changed them to unique variables before we applied \textsc{GeNME}.

With the positive example \ps{arch(struct1)} and the rewriting filters\\ $\vmapping{\ps{supports}}{\ps{supported\_by}}$, $\vmapping{\ps{supported\_by}}{\ps{supports}}$, $\vmapping{\ps{meets}}{\ps{not\_meets}}$, and\\ $\vmapping{\ps{not\_meets}}{\ps{meets}}$ that change all occurrences, \textsc{GeNME} yields the following near miss explanation with degree $d = 1$:

\begin{center}
$
\begin{aligned}
    \ps{arch}(struct4) \gets & \ps{contains}(struct4, a1), \ps{contains}(struct4, b), \ps{contains}(struct4, c),\\
    & \ps{is\_a}(a1, wedge), \ps{is\_a}(b, brick), \ps{is\_a}(c, brick),\\
    & \ps{supports}(b, a1, struct4), \ps{supports}(c, a1, struct4), \\
    & \ps{meets}(b, c, struct4)
\end{aligned}
$
\end{center}

$\nmx \ps{arch}(struct4)$ is in fact a plausible near miss example since the only feature that has changed is the closure of the passage between the pillars. The wedge roof is a feature from the positive example that also holds for the miss.

Other explanations are found for $d = 3$:

\begin{center}
$
\begin{aligned}
    \ps{arch}(struct6) \gets & \ps{contains}(struct6, a2), \ps{contains}(struct6, b), \ps{contains}(struct6, c),\\
    & \ps{is\_a}(a2, brick), \ps{is\_a}(b, brick), \ps{is\_a}(c, brick),\\
    & \ps{supports}(b, a2, struct6), \ps{supports}(c, a2, struct6), \\
    & \ps{meets}(b, c, struct6)
\end{aligned}
$
\end{center}

\begin{center}
$
\begin{aligned}
    \ps{arch}(struct5) \gets & \ps{contains}(struct5, a2), \ps{contains}(struct5, b), \ps{contains}(struct5, c),\\
    & \ps{is\_a}(a2, brick), \ps{is\_a}(b, brick), \ps{is\_a}(c, brick),\\
    & \ps{supported\_by}(b, a2, struct5), \ps{supported\_by}(c, a2, struct5), \\
    & \ps{not\_meets}(b, c, struct5)
\end{aligned}
$
\end{center}

The former explanation features the far miss example $\nmx\ps{arch}(struct6)$ where not only the two pillars meet, but also the roof got a new shape. The latter explanation deals with the fact that $\nmx\ps{arch}(struct5)$ turns around the \ps{supports} relation between pillars and roof. Both miss examples are farther away from the used positive example and might not be that helpful in explaining what an arch is.

%booktabs (toprule, midrule, bottomrule)
\begin{table}[t]
	\centering
	\caption{Number of found near miss explanations by degree in the Winston arches domain. $\mathcal{N}$ denotes the set of all near miss candidates, $\mathcal{E}_d$ the set of near miss explanations of degree $d$. $\bimapping{x}{y}$ denotes the use of $\mathcal{V}_{x \mapsto y}$ or $\mathcal{V}_{y \mapsto x}$, respectively.}
	\label{tab:nme_arches}
	\begin{tabular}{cc}
		& $\ps{arch}(struct1)$\\
		& $\lvert\mathcal{N}\rvert = 3$\\
		\toprule
		$\bimapping{meets}{not\_meets}$ &\\
		$\lvert \mathcal{E}_1 \rvert$ & 1\\
		$\lvert \mathcal{E}_2 \rvert$ & 0\\
		$\lvert \mathcal{E}_3 \rvert$ & 1\\
		\midrule
		$\bimapping{supports}{supported\_by}$ &\\
		$\lvert \mathcal{E}_1 \rvert$ & 0\\
		$\lvert \mathcal{E}_2 \rvert$ & 0\\
		$\lvert \mathcal{E}_3 \rvert$ & 1\\
		\bottomrule
	\end{tabular}
\end{table}

Table~\ref{tab:nme_arches} shows the number of near misses \textsc{GeNME} found for the different rewriting filters.

\subsection{File Management}

\begin{figure}[t]
	\centering
	\textbf{Background Knowledge:}\\
	\begin{tabular}{l}
		\texttt{file(file10)} \\
		\texttt{file\_name('1fTmw4WN.PNG', file10)} \\
		\texttt{media\_type(png, file10)} \\
		\texttt{file\_size(6902, file10)} \\
		\texttt{creation\_time('1984-12-18', file10)}\\
		\\
		\texttt{file(file11)} \\
		\texttt{file\_name('Sv4Xy5n6.PNG', file11)} \\
		\texttt{media\_type(png, file11)} \\
		\texttt{file\_size(12287, file11)} \\
		\texttt{creation\_time('1996-12-20', file11)} \\
	\end{tabular}
	
	\textbf{Selection of positive examples:}\\
	\texttt{irrelevant(file10)}\\
	\texttt{irrelevant(file11)}
	
	\textbf{Selection of negative examples:}\\
	\texttt{irrelevant(file121)}\\
	\texttt{irrelevant(file168)}
	\caption{Excerpt of background knowledge for the file management domain together with a selection of positive and negative examples for the \ps{irrelevant} concept.}
	\label{fig:dare2del_ilp}
\end{figure}

The purpose of the file management domain is to identify irrelevant files, that is files that may be deleted by the user (see~\citep{SiebersS19}). For this, a file system with related files and folders is modeled. The domain is represented by relations such as\\
\ps{creation\_time}, \ps{file\_size}, \ps{file\_name}, and \ps{media\_type}. Figure~\ref{fig:dare2del_ilp} shows an excerpt of the background clauses of the theory for a randomly generated file system. Additionally, the theory contains clauses for several auxiliary relations, such as \ps{older}, \ps{larger}, and \ps{in\_same\_folder}. To facilitate matters, we assume that the learned irrelevancy concept is formalized in a single clause (see Equation~\ref{eq:irrelevant}).

\begin{equation}
\begin{aligned}
\ps{irrelevant}(F) \gets & \ps{in\_same\_folder}(F, G), \\
& \ps{media\_type}(M, F), \\
& \ps{media\_type}(M, G), \\
& \ps{older}(F, G) \label{eq:irrelevant}
\end{aligned}
\end{equation}

We use \textsc{GeNME} to explain two arbitrary positive examples, \\$\ps{irrelevant}(file10)$ and $\ps{irrelevant}(file112)$. As for the family domain, we provide rewriting filters $\vmapping{\ps{older}}{\ps{newer}}$ and $\vmapping{\ps{newer}}{\ps{older}}$ where both allow changing a single literal.

\begin{table}[t]
	\centering
	\caption{Number of found near miss explanations by degree in the file management domain. $\mathcal{N}$ denotes the set of all near miss candidates, $\mathcal{E}_d$ the set of near miss explanations of degree $d$ and \ps{irr} the \ps{irrelevant} concept. $\bimapping{x}{y}$ denotes the use of $\mathcal{V}_{x \mapsto y}$ or $\mathcal{V}_{y \mapsto x}$, respectively.}
	\label{tab:dare2del}
	\begin{tabular}{ccc}
		& $\ps{irr}(file10)$ & $\ps{irr}(file112)$\\
		& $\lvert\mathcal{N}\rvert = 80$ & $\lvert\mathcal{N}\rvert = 80$\\
		\toprule
		$\bimapping{older}{newer}$ &  &\\
		$\lvert \mathcal{E}_1 \rvert$ & 1 & 1 \\
		$\lvert \mathcal{E}_2 \rvert$ & 19 & 8 \\
		$\lvert \mathcal{E}_3 \rvert$ & 48 & 59 \\
		\bottomrule
	\end{tabular}
\end{table}

As shown in Table~\ref{tab:dare2del}, \textsc{GeNME} identifies 68 near miss candidates as near miss examples for both examples (85.0\,\%). Nevertheless, a single near miss example is identified as nearest example (degree 1) for each example. Both are files of the same media type in the same folder which are newer than the file from the example.

\section{Empirical Study of Human Preferences of Explanation Types}

To investigate whether near miss explanations are considered useful by humans, we conducted an empirical study on preferences of explanation modalities for the abstract relational family domain and the visual relational arches domain. For both domains, a cover story has been given and participants had to evaluate the helpfulness of explanations given this setting which is described in the following subsections.

Overall, four different types of explanations have been considered:
\begin{itemize}
    \item \textbf{General rule (R):} a global explanation of the concept a specific instance belongs to,
    \item \textbf{Example (E):} an example-based explanation in form of a specific instance belonging to the concept,
    \item \textbf{Near Miss (N):} a contrastive example which has a high degree of structural similarity to the specific instance under consideration,
    \item \textbf{Far Miss (F):} a negative example for the considered concept which has a low degree of structural similarity to the specific instance under consideration.
\end{itemize}

\noindent
All explanations were presented in form of natural language sentences. Such natural language explanations can be generated from ILP learned rules in a straight-forwards manner \citep{SiebersS19}. 

These four types of explanations address different information needs \citep{Miller19}: To understand the general concept, a global rule can be assumed to be especially helpful. However, it might be the case that the helpfulness is different for abstract in contrast to visual domains. In the second case, a visual prototype might be more effective \citep{gurumoorthy2019efficient}. In cognitive psychology, visual prototypes have been shown to be an effective means of concept representation for basic categories \citep{rosch1978principles}. For simple domains, an arbitrary instance might convey information similar to a prototype. For instance, in medical text books, example images are given to illustrate what a specific skin disease looks like. A near miss example should be especially helpful to highlight what (missing) information would be necessary to make an object belong to a class \citep{gentner1994structural}. This is often helpful, if feature values or relations are hard to grasp. For instance, mushroom pickers use images to distinguish an edible mushroom from the visually most similar toadstool. Arbitrary negative examples, especially far misses can be assumed to be less helpful to understand a concept or why a specific instance belongs to a concept. This type of explanation has been introduced as a baseline.
We assume that combining different types of explanations can be more effective than each of these explanations alone. Especially, a combination of a global rule with a near miss might be most efficient to explain relational concepts.

Given the cover story, the helpfulness of the different types of explanations has been assessed with a complete pairwise comparisons \citep{thurstone1927law}. In a first part of the study, all pairings of the four explanation types have been presented in a arbitrary sequence and participants had to evaluate which they found more helpful as an explanation given the cover story. In a second part of the study, pairs of pairs of explanations have been presented. In a final part, the helpfulness of the four explanation types for different information needs as been assessed explicitly. Participant rated how helpful an explanation is to understand 
\begin{itemize}
    \item the \textbf{general} concept,
    \item a particular \textbf{example} instance for the concept,
    \item what is \emph{not} in the concept (\textbf{exclusion}).
\end{itemize}
\noindent
on a scale from 1 to 5 with labels \emph{not at all} to \emph{absolutely}.

The study was conducted as an online experiment with 73 participants (42 females, 31 males) with average age 35.72 (min 18, max 64). 43 participants were employed, 27 where students, 2 were self-employed and 1 person was retired. About 50\% of the participants received first the family domain followed by the arches domain and the other half of participants started with the arches followed by the family domain.

%The age was gathered in categories and the average age of the participants can therefore be given as between 33.42 and 37.01. %TODO comment on age and standard deviation
%TODO Age was assessed in bins of...
%Participants were presented with two different situations (see domains below) where they should explain a given relational concept. For each of the situations they were then presented with pairings of explanations where they had to state their preference as to which explanation they wanted to use to explain the concept. The possible base explanations were a natural language rule stating the concept (R), an example instance from the concept (E), a miss explanation of low degree coming from \textsc{GeNME} (N) and a miss explanation of high degree coming from \textsc{GeNME} (F). For the rest of this chapter we will refer to misses with low degree generally as \emph{near misses} and for misses with high degree as \emph{far misses}. We also converted the miss explanations to natural language. The first set of decision pairings was designed as a shuffled complete pairwise comparison of R, E, N and F resulting in six pairings. The second set was also designed as a shuffled complete pairwise comparison between pairs of the base explanations resulting in 15 pairings. The order inside the pairs follows the ranking R before E before N before F.

%Hypotheses
Although this is an exploratory study, given the considerations above, we can formulate the following hypotheses:  (1) Near miss examples should be preferred over far miss examples in the pairwise comparisons; (2) Near miss examples should be rated as the most helpful to understand the boundaries of a given concept; (3) For the visual domain example-based visual explanations should be preferred over abstract (verbal) rule-based explanations.

%Our hypothesis is that explanation modalities featuring near miss explanations resulting from \textsc{GeNME} are preferred by humans over far misses to understand a given learned relational concept. We also expect near misses to be helpful in understanding the boundaries of a given concept (Which examples are excluded from the concept). Further we assume that the preference also changes depending on the given domain. While rule based explanations should be perceived as more helpful in abstract domains, example based explanations should be preferred in visual domains. In the following we state the results for the two domains:
%TODO hier müsste die Argumentation mit Rosch, 1999 hin?
%TODO also, when having a visual domain where examples are more important than rules, a near miss explanation is preferred over a positive example?

\subsection{Results for the Family Domain}

The family domain (see Sections~\ref{sec:ilp} and~\ref{sec:app_family}) has been presented with the family tree of Kate as given in Figure~\ref{fig:familytree} with regular arrows and a legend that does not highlight miss examples. Participants were asked to imagine a conversation with their friend Kate who is originating from a native American tribe. She is curious about the different definitions of family relations in western culture, since she is not familiar with them and the definitions that she grew up with are very different from the participant's. In particular, she wants to understand the \textbf{grandfather} relation between Kate and Ian.

The participants were then asked to give their preference on explanations in the pattern outlined above: 

\begin{itemize}
    \item (R) A grandfather is a male parent of one of your parents.
    \item (E) One of your parents, Tom, has a male parent called Ian. Ian is your grandfather.
    \item (N) Jodie, the female parent of your parent Tom is NOT your grandfather; it is your grandmother.
    \item (F) Mat, the male child of Tom, who is the child of Ian is NOT the grandfather of Ian; it is his grandson.
\end{itemize}

For the six simple explanation pairings we observed the following frequency ranking (rounded relative frequencies in brackets): R (0.43) $>$ E (0.37) $>$ N (0.17) $>$ F (0.03). The rule for the relational concept was preferred over all other explanation modalities followed by an example and a miss with lowest degree. We see that for an abstract relational domain, a rule describing the concepts seems to be more helpful than stating an example. Nevertheless when having the choice between a miss example closer to the chosen example or a miss example further away, the participants preferred the former one confirming the hypothesis that the choice of miss examples is important.

For the second set of 15 the pairwise comparison of pairings of the base explanations we observed the frequency ranking RE (0.32) $>$ RN (0.21) $>$ EN (0.19) $>$ EF (0.13) $>$ RF (0.13) $>$ NF (0.02). The rule-example pairing is favored over all other pairings followed by the rule-low degree miss explanation pairing. Again, a preference for the rule shows in this abstract domain.

\begin{table}[t]
	\centering
	\caption{Results of the questions on which explanations fulfilled which purpose in the family domain. For each purpose (rows) the mean rating value over all participants is given for each explanation. Bold numbers highlight the highest value for each purpose.}
	\label{tab:decisions_family}
	\begin{tabular}{c|cccc}
		 & (R)ule & (E)xample & (N)ear Miss & (F)ar Miss \\
		\toprule
		\emph{general} & \textbf{4.97} & 4.52 & 2.93 & 2.19 \\
		\emph{example} & 4.14 & \textbf{4.70} & 2.49 & 2.37 \\
		\emph{exclusion} & 2.95 & 2.62 & \textbf{4.30} & 3.67 \\
		\bottomrule
	\end{tabular}
\end{table}

Table~\ref{tab:decisions_family} shows the results of the questions on purpose of explanations. The near miss explanation, although preferred over the far miss explanation, is rather unsuited to understand the general concept of this abstract relational domain as well as understanding why a particular example belongs to the concept. For the purpose of understanding the boundaries of the concept however, the near miss explanation performs best compared to all other explanations. We therefore postulate that a near miss explanation can help humans to avoid false positives when making decisions.

\subsection{Results for the Arches Domain}

The Winston arches domain has been introduced as shown in Figure~\ref{fig:posneg_arches} without the object labels and with a focus on the arch labeled \ps{struct1}. Participants were asked to imagine playing with building blocks with their five year old son. They want to show him some new structure called \textbf{arch} given the examples and counterexamples in Figure~\ref{fig:posneg_arches}.

The explanations for the arch domain are:

\begin{itemize}
    \item (R) An arch consists of two rectangle blocks that do not touch. They support a triangle block.
    \item (E) \emph{We showed the structure labeled \ps{struct1} as a positive example.}
    \item (N) \emph{We showed the structure labeled \ps{struct4} as a near miss.}
    \item (F) \emph{We showed the structure labeled \ps{struct6} as a far miss.}
\end{itemize}

The first set of six questions yields the frequency ranking E (0.45) $>$ R (0.30) $>$ N (0.18) $>$ F (0.08). For this visual domain it comes with no surprise that examples seem to be more helpful for explaining the concept. Also, the near miss explanation is preferred over the far miss explanation.

The 15 pairing-decisions yield the ranking EN (0.27) $>$ RE (0.25) $>$ EF (0.21) $>$ RN (0.14) $>$ RF (0.08) $>$ NF (0.04). This time the explanation pair example-near miss is a clear favorite. It seems that humans find stating an example along with a near miss more helpful for explaining concepts in a visual domain than to only state a positive example along with the rule that describe the concept. Also, a rule in combination with a near miss seems to be more helpful than a rule with a miss further away.

\begin{table}[t]
	\centering
	\caption{Results of the questions on which explanations fulfilled which purpose in the arches domain. For each purpose (rows) the mean rating value over all participants is given for each explanation. Bold numbers highlight the highest value for each purpose.}
	\label{tab:decisions_arches}
	\begin{tabular}{c|cccc}
		 & (R)ule & (E)xample & (N)ear Miss & (F)ar Miss \\
		\toprule
		\emph{general} & 4.45 & \textbf{4.70} & 2.70 & 2.36 \\
		\emph{example} & \textbf{4.56} & 4.27 & 2.74 & 2.38 \\
		\emph{exclusion} & 3.25 & 2.73 & \textbf{3.95} & 3.82 \\
		\bottomrule
	\end{tabular}
\end{table}

Table~\ref{tab:decisions_arches} states the results for the questions on purpose for the arches domain. As expected, the statement of an example seems to help humans better in understanding the visual concept than stating a general rule. Surprisingly the rule was better suited to understand why a particular example belongs to the concept.
Both near and far miss were better suited to understand what is not in the concept and therefore can be used to sharpen the boundaries of the given concept.

\subsection{Discussion of Empirical Results}

Given the presented results, the hypothesis that near miss examples should be preferred over far miss examples in the pairwise comparisons has been confirmed with one exception: For the arches domain the combination of example and far miss is rated as being more helpful as the combination of rule and near miss explanation. In accordance with the observation that examples are especially helpful as explanations in visual domains, a positive example together with a negative example for the concept of an arch is an effective means to indicate what aspects are necessary for a structure to belong to the concept.

Our hypothesis that near miss examples should be rated as the most helpful to understand the boundaries of a given concept has been confirmed for the family as well as for the arches domain. In the arches domain the difference between the helpfulness ratings of near vs. far miss explanations are not very pronounced while for the family domain near miss explanations have been rated as much more helpful than far misses. This finding might be due to the rather simple category structure of the arches domain where arbitrary negative examples might be suitable to highlight the relevance of what is missing to belong to the concept.

The third hypothesis has been confirmed for the helpfulness ratings for the visual domain. The example has been rated as more helpful as the rule to explain the general concept. Furthermore the example has been preferred over the rule in the complete pairwise comparison and the combination of example and near miss example has been preferred most often.

\section{Conclusions and Further Work}

We introduced near miss explanations for relational concepts learned with ILP. The \textsc{GeNME} algorithm has been presented which generates contrastive examples with different degrees of nearness to a specific positive instance for a given concept. The explanations facilitate identifying crucial aspects of the concept. The current version of \textsc{GeNME} relies on checking all candidate examples which is not feasible for domains with a large number of constants. There, it might be necessary to restrict candidates to candidates already \emph{close} to the positive example. One way to achieve this might be considering similarities within the graph spanned by the constants and predicate relations in the domain.

In an empirical study we could demonstrate that human rate near miss examples as helpful, especially in combination with a global rules-based explanation and that humans evaluate near misses as especially helpful to understand the crucial aspects for instances which make them belong to a specific concept. 

At the beginning, focus of research on explainable AI (XAI) has been on visual highlighting of relevant information for blackbox classifiers which is mostly of interest for the model developers \citep{samek2019explainable}. Recently, there is a growing interest of explanations for domain experts and end-users \citep{arrieta2020explainable,schmid2021interactive}. For these groups, it is of special importance that explanations are easily comprehensible and effectively bring across the most helpful information for a given task. Therefore, the next challenge of XAI systems will be to consider explanations of different modality and provide strategies to select the probably most helpful explanation for a given context \citet{williams2010role}. As cognitive science research suggests \citep{gentner1994structural}, near miss explanations can play an important role to highlight what aspects are necessary for an instance to belong to a given class.

\section*{Acknowledgements}
We thank Sebastian Seufert and Klaus Stein for support with the generation of the of the domains and for helpful discussions of the near miss algorithm.

\bibliography{samplepaper}

\end{document}